\documentclass[11pt]{article}
\usepackage[margin=1in]{geometry}
\usepackage{amsmath}
\usepackage{amssymb}
\usepackage{booktabs}
\usepackage{graphicx}
\usepackage[hidelinks]{hyperref}
\usepackage{microtype}

\title{Self-Supervised Pretext Tasks for Infant Cry Analysis:\\
A Controlled Comparison and a Cautionary Result on Donateacry}
\author{Luigi Simeone\\ \small Independent researcher}
\date{August 2026}

\begin{document}
\maketitle

\begin{abstract}
We compare six self-supervised pretext tasks for infant cry analysis under
a fixed budget, meaning the same compact encoder of 1.17M parameters, the
same 115 hours of license-verified public pretraining audio, and the same
evaluation protocol for every candidate. On cry detection the
reconstructive objectives dominate, and a linear probe over a
masked-spectrogram encoder reaches 0.988 AUC with subject-wise splits even
though the encoder never observed a cry during pretraining. On cry-reason
classification over donateacry, the de facto public benchmark for cry
reasons, every encoder performs at chance (0.38 to 0.54 macro AUC over 5
classes), and neither domain adaptation on 1.8 hours of real cries nor
end-to-end fine-tuning moves the result. Since a frozen HuBERT-base with
80 times more parameters shows the same pattern, the bottleneck must sit
in the labels and not in model capacity. We then reproduce the 90\%+
accuracies of the donateacry literature on our own system by changing
nothing but the evaluation protocol: clip-wise splits raise accuracy to
85.2\% (barely above the 83.8\% majority-class baseline), and applying
augmentation before splitting raises it to 97.9\%, matching the reported
state of the art, from the same model that measures 0.49 macro AUC under
subject-wise splits. Under leakage-free splits, a twentyfold augmentation
of the labeled set (vocoder speaker perturbation and noise mixing, 21
hours) leaves cross-subject AUC unchanged: for this task the effective
sample size is the number of infants. We release code, seeds and per-clip license
manifests.\footnote{\url{https://github.com/PhysicsInforMe/infant-cry-ssl}}
\end{abstract}

\section{Introduction}

Infant cry analysis has two natural tasks: detection asks whether a sound
is an infant crying, and reason classification asks why. The strongest
published result on reason classification comes from Gorin et
al.~\cite{gorin2023}, who reach 74.5 macro AUC on three cry triggers with
self-supervised pretraining, cry-domain adaptation and an 80M-parameter
CNN14, on a private clinical corpus of about 1{,}150 recordings with
trigger labels assigned by medical and research staff. Around that result
sits a larger body of work reporting 90\%+ accuracy on donateacry, a
public corpus of 457 parent-labeled clips. The two numbers are not even
the same metric (macro AUC against unbalanced accuracy), and the gap
between a hard-won 74.5 AUC on clinical data and near-perfect accuracy on
a small volunteer corpus deserves suspicion. This paper supplies the
explanation with measurements.

We ask three questions. First, do the conclusions of \cite{gorin2023}
survive a move to public data and a compact encoder sized for edge
deployment? Second, which pretext task earns its keep at that scale? SSL
work on cries has so far committed to a single objective per paper;
we compare six under identical conditions, including two variants of a
hypothesis about teaching frequency-band structure explicitly, designed so
that the failure modes we predicted in advance would show up as numbers
instead of remaining a matter of opinion. Third, what do donateacry's
labels actually support once
evaluation respects subject identity everywhere, including inside the SSL
stages?

Contributions: (i) a controlled six-task pretext comparison at fixed
budget, with reconstructive objectives clearly ahead for cry detection;
(ii) a negative result on donateacry reason classification that survives a
model-capacity control, together with an end-to-end reproduction of the
literature's 90\%+ accuracies obtained purely by adopting its evaluation
protocol; (iii) a leakage-free protocol for evaluating SSL
domain adaptation, with adaptation re-run per fold on training subjects
only; (iv) evidence from an augmentation ablation that labeled subjects,
not labeled clips, are the binding resource; (v) an ear-verified
hard-negative set, including adult cries mislabeled as infant cries in
FSD50K ground truth.

\section{Related work}

\paragraph{Infant cry analysis.} The field predates deep learning by
decades: spectrographic studies in the 1960s already cataloged cry types
and their acoustic correlates~\cite{waszhockert1968}, and later clinical
reviews treated caregiver perception of cries as a perceptual process to
be studied in its own right, alongside acoustic
analysis~\cite{lagasse2005}, rather than as ground truth, a distinction
our results land on hard. The
machine-learning line runs through pathology detection on the small Baby
Chillanto corpus~\cite{reyes2004} to the Ubenwa group's work on
cry-based screening of perinatal asphyxia with transfer
learning~\cite{onu2019}, the CryCeleb verification
benchmark~\cite{cryceleb2023}, and the SSL study we take as our
reference~\cite{gorin2023}: CNN14~\cite{kong2020} pretrained with
SimCLR-style contrastive learning~\cite{simclr}, adapted on 11 hours of
unlabeled cries, fine-tuned on staff-assigned trigger labels, evaluated
with per-patient splits. CryCeleb matters to us for a second reason: it
demonstrates that infant identity is readily decodable from cry acoustics,
which is exactly the signal a clip-level split hands to a classifier for
free.

\paragraph{Reason classification on donateacry.} A separate body of work
trains reason classifiers directly on donateacry and reports accuracies in
the mid-90s. The strongest recent example~\cite{frontiers2024} reaches
96.4\% with MFCC features and a random forest, improving on a
scalogram-based system it cites at 95.2\%; its split is described only as
80/20, with no mention of subject identity. We are not aware of any
donateacry result that reports subject-wise evaluation, which is the gap
this paper fills.

\paragraph{Evaluation leakage in clinical machine learning.} The failure
mode is well documented outside audio: Saeb et al.~\cite{saeb2017} showed
that record-wise cross-validation on clinical sensor data inflates
accuracy relative to subject-wise evaluation, and argued that a split
should approximate the use case, because a deployed model meets new
patients and never new records of patients it already knows. Kapoor and Narayanan~\cite{kapoor2023}
survey leakage across ML-based science and identify it as a leading cause
of irreproducible results. Our contribution to this literature is a
domain-specific, fully controlled instance: same model, same data, three
protocols, with the invalid ones reproducing the published numbers.

\paragraph{Self-supervised audio representations.} The pretext tasks we
compare are compact instances of the field's main families: contrastive
and distillation-style joint embedding~\cite{simclr,simsiam}, with
BYOL-A~\cite{byola} the closest precedent for small-encoder audio SSL;
masked reconstruction, from masked autoencoders in
vision~\cite{mae} to Audio-MAE~\cite{audiomae}; and masked prediction of
latent targets, the family of wav2vec~2.0~\cite{wav2vec2} and
HuBERT~\cite{hubert}, which enters our study only as a frozen capacity
control. FSD50K~\cite{fsd50k} and VocalSound~\cite{vocalsound} supply the
pretraining pool. What our comparison adds to this literature is a
controlled ranking at a scale, about one million parameters, that the
benchmark papers rarely visit.

\section{Data}

All corpora are public and were filtered clip by clip for permissive
licenses, with the admitted list and attributions generated by the pipeline
itself: FSD50K (43{,}379 clips, 90.1 h, CC0/CC BY, with 7{,}818 CC-BY-NC
clips excluded), VocalSound (20{,}985 clips, 24.4 h, CC BY-SA), the
cleaned release of donateacry~\cite{donateacry} (447 clips, 0.9 h,
ODbL/DbCL), and 204 clips (2.1 h) retrieved from
Freesound by direct query and deduplicated against FSD50K. Audio is cached
as memory-mapped float16 log-mel with 64 bands at 16 kHz (25 ms window,
10 ms hop).

Labels received a human pass. A PANNs Cnn14 tagger~\cite{kong2020} triaged
the pool; 127
borderline cases, selected by stratified value of information, were then
labeled by ear, with a second pass separating infant cries from adult cries
and animal sounds. Two outcomes matter downstream. About half of the
FSD50K clips tagged \emph{Baby cry, infant cry} that survived triage are
adults crying; they enter the evaluation as hard negatives. And 11 of the
15 lowest-scoring donateacry ``cries'' are not cries, so 14 spurious clips
were excluded from all supervised experiments, leaving 433 clips from 204
infants.

\section{Method}

\begin{figure}[t]
\centering
\includegraphics[width=\linewidth]{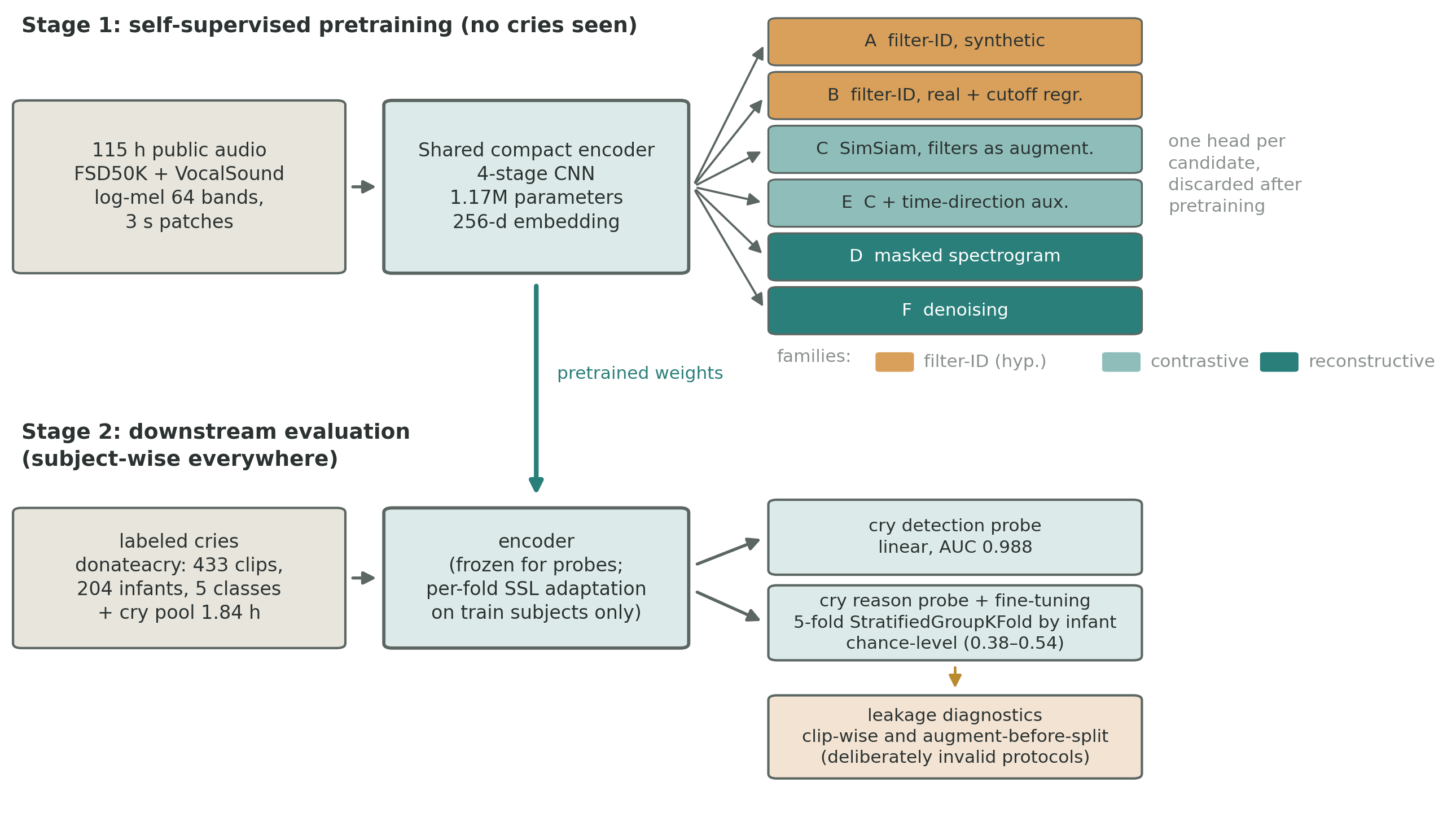}
\caption{Study design. Stage 1: one compact encoder is pretrained six
times, once per pretext task, on 115 hours of public audio that contains
no infant cries; the task heads are discarded. Colors group the tasks
into three families. Stage 2: the pretrained encoder is evaluated on
labeled cries with subject-wise splits everywhere, including inside the
per-fold SSL adaptation; the deliberately invalid protocols at the bottom
exist only to measure leakage.}
\label{fig:arch}
\end{figure}

Figure~\ref{fig:arch} gives the overview of the design, in which a single
encoder feeds six interchangeable pretext heads during pretraining and one
shared evaluation harness downstream; the rest of this section describes
each piece in turn.

\paragraph{Encoder.} A four-stage CNN (two $3{\times}3$ convolutions per
stage, BatchNorm, stride-2 downsampling), global average pooling and a
256-d embedding: 1.17M parameters. Every candidate shares this encoder and
differs only in a head that is discarded after pretraining. Inputs are
3-second patches (300 frames $\times$ 64 bands), normalized per clip.

\paragraph{Pretext tasks.} (A) \emph{Filter-ID on synthetic signals}:
classify which of four band filters (low-pass, high-pass, band-pass,
band-stop) was applied to synthetic harmonic spectra with F0 drawn from the
infant range and slow amplitude modulation. (B) \emph{Filter-ID on real
audio}: the same four-way classification on real patches, plus continuous
regression of the cutoff bands. (C) \emph{SimSiam} with the band filters
demoted to augmentations, alongside noise mixing with other batch samples,
gain jitter, small pitch and time shifts, and time masking.
(D) \emph{Masked spectrogram modeling}, reconstructing masked
time-frequency blocks through a light transposed-convolution decoder,
trained from scratch. (E) Candidate C plus an auxiliary time-direction
classifier. (F) \emph{Denoising}: reconstruct the clean log-mel from a
mix with another sample plus gain jitter.

\paragraph{Why these six.} The set was assembled so that adjacent
candidates differ by a single ingredient, which turns the final ranking
into a series of controlled answers where a survey of popular objectives
would only have produced a leaderboard. The starting point is a domain
hypothesis we chose to test
head-on: infant cries carry their information in a known frequency
structure (fundamental at 250 to 700 Hz, harmonics above), so a pretext
that forces the encoder to identify band filters should teach frequency
awareness cheaply, with labels that cost nothing. Candidates A and B are
that hypothesis in its pure and strengthened forms, and three weaknesses
were written down before training: the task can be solved from band-energy
statistics alone, the way rotation prediction collapsed in vision; it is
static, so it never rewards temporal sensitivity, while the class-relevant
structure of cries (rhythm, melodic contour) is temporal; and synthetic
stationary signals lack the natural statistics SSL feeds on.

Each remaining comparison isolates one question. Moving the same task
from synthetic to real audio (A against B) asks whether natural statistics
recover the transfer that synthesis loses. Demoting the identical filters
from labels to distortions inside SimSiam (B against C) asks what role
hand-designed acoustic knowledge should play in the objective at all, and
C is in that sense the built-in counter-hypothesis. Attaching an auxiliary
time-direction task (C against E) asks whether forced temporal sensitivity
contributes anything an invariance objective misses, while swapping masked
blocks for realistic noise (D against F) asks whether the corruption type
matters inside the reconstructive family. The widest question,
reconstruction against invariance, falls out of comparing the D/F pair
with C/E at a scale of 1.17M parameters and batch 256, where the
literature offers little guidance because most SSL results are reported
far larger. The bake-off puts numbers on every one of these edges at
once.

\paragraph{The filter-ID objectives, formally.} To our knowledge,
band-filter identification has not previously been used as a pretext task
in audio SSL; band manipulation appears in the literature only as an
augmentation, most prominently the frequency masking of
SpecAugment~\cite{specaugment}, and candidate C is exactly that usage.
Since A and B are new, we state them precisely. Let
$x \in \mathbb{R}^{T\times M}$ be a per-clip-normalized log-mel patch with
$T{=}300$ frames and $M{=}64$ bands. The filter acts additively in the log
domain,
\begin{equation}
(\mathcal{T}_{t,a,b,\gamma}\,x)_{\tau m} \;=\; x_{\tau m} \;-\;
\gamma\, S_t(m;\, a, b),
\end{equation}
where $\gamma \sim \mathcal{U}(3,6)$ is the attenuation depth and
$S_t(\cdot) \in \{0,1\}$ marks the stop region of filter type
$t \in \{\text{LP, HP, BP, BS}\}$: bands above $a$, below $a$, outside
$[a,b]$, or inside $[a,b]$ respectively. Candidate A draws its inputs from
a synthetic generator instead of the corpus: a harmonic template with
fundamental $f_0 \sim \mathcal{U}(250, 700)$~Hz placed at the mel
positions of $k f_0$ with partial amplitudes decaying as $k^{-1/2}$, a
constant noise floor, and a slow sinusoidal amplitude modulation at
$f_{\mathrm{am}} \sim \mathcal{U}(1, 8)$~Hz, so that the signal carries
the gross spectral layout of a cry while carrying none of its natural
statistics. Its loss is cross-entropy on the filter type,
\begin{equation}
\mathcal{L}_A \;=\; \mathrm{CE}\!\big(h_A(g(\mathcal{T}x)),\; t\big),
\end{equation}
with $g$ the shared encoder and $h_A$ a linear head. Candidate B applies
$\mathcal{T}$ to real patches and adds continuous recovery of the
cutoffs,
\begin{equation}
\mathcal{L}_B \;=\; \mathrm{CE}\!\big(h_{\mathrm{type}}(g(\mathcal{T}x)),\; t\big)
\;+\; \Big\lVert \sigma\!\big(h_{\mathrm{cut}}(g(\mathcal{T}x))\big) -
\tfrac{1}{M-1}(a, b) \Big\rVert_2^2,
\end{equation}
which removes the four-way shortcut and forces a finer reading of the
spectrum than type classification alone. The remaining objectives are
standard and we only fix their instantiation: the symmetric negative
cosine of SimSiam for C and E~\cite{simsiam}, with E adding a binary
cross-entropy on time direction weighted at 0.2, and mean squared error
in input space, on masked cells for D and on the full clean target for F.

\paragraph{Budget.} Each candidate trains for 8{,}000 steps at batch 256
(AdamW, cosine schedule, mixed precision) on FSD50K+VocalSound only.
Donateacry and the Freesound cries are held out of pretraining entirely so
that downstream probes stay uncontaminated. One run takes under an hour on
a laptop RTX 4060.

\paragraph{Evaluation.} All splits are by subject: the contributor prefix
for donateacry, the uploader for Freesound. StratifiedGroupKFold with 5
folds; macro one-vs-rest AUC, balanced accuracy, ECE. The rule extends to
the SSL stages: when we adapt an encoder on cry audio, adaptation is re-run
for every fold using only that fold's training subjects. The reason is
concrete: an encoder adapted on a given infant's audio embeds that
infant's voice characteristics, so probing it afterwards on the same
infant partly measures recognition of a voice it has already seen, which
is leakage at the representation level even though no label was ever
touched. The cost of doing this properly is five adaptations per
candidate instead of one, and at this scale that means minutes per fold. A clip-wise split is computed once,
deliberately, as a diagnostic of what leakage buys.

\paragraph{Augmentation ablation.} Each donateacry clip is expanded with
PSOLA vocoder variants (F0 shifts of 0.5 to 3 semitones, formant scaling
0.88 to 1.12, timing preserved) and with mixes against domestic noise from
the license-clean FSD50K pool at 3 to 20 dB SNR, 21 hours in total,
class-balanced. Synthetic variants inherit the source infant's identity
for splitting; evaluation uses real audio only.

\section{Results}

\begin{table}[t]
\centering
\caption{Cry/non-cry linear probe, subject-wise 5-fold (541 verified
positives, 428 negatives of which 30 hard).}
\label{tab:detect}
\begin{tabular}{lc}
\toprule
Pretext & AUC \\
\midrule
D\quad masked spectrogram & $0.988 \pm 0.004$ \\
F\quad denoising & $0.982 \pm 0.006$ \\
C\quad SimSiam + filter augmentation & $0.972 \pm 0.015$ \\
E\quad C + temporal auxiliary & $0.967 \pm 0.020$ \\
B\quad filter-ID (real) & $0.880 \pm 0.024$ \\
A\quad filter-ID (synthetic) & $0.793 \pm 0.040$ \\
\bottomrule
\end{tabular}
\end{table}

\begin{table}[t]
\centering
\caption{Cry-reason classification on donateacry (5 classes, 433 clips,
204 infants, subject-wise unless stated).}
\label{tab:reason}
\begin{tabular}{lc}
\toprule
Setting & Macro AUC \\
\midrule
Best pretext, linear probe (E) & $0.539 \pm 0.034$ \\
Other pretexts, linear probe & 0.38--0.47 \\
After per-fold SSL adaptation (1.84 h of cries) & 0.33--0.45 \\
Frozen HuBERT-base, 94M params (control) & 0.42 \\
Same embeddings, clip-wise split (diagnostic) & 0.58--0.61 \\
End-to-end fine-tuning, best of 4 augmentation arms & $0.49 \pm 0.11$ \\
\midrule
\multicolumn{2}{l}{\emph{Deliberately leaky, replicating the 90\%+ literature:}} \\
End-to-end, clip-wise split & 0.70 \ (85.2\% acc.) \\
End-to-end, clip-wise + augment-before-split & 0.976 \ (97.9\% acc.) \\
Majority-class baseline (``always hungry'') & 83.8\% acc. \\
\midrule
Gorin et al.~\cite{gorin2023}, clinical labels, 3 classes & 74.5 \\
\bottomrule
\end{tabular}
\end{table}

\paragraph{Detection.} Table~\ref{tab:detect} settles the pretext
question at this scale. Reconstructive objectives transfer best, the
contrastive pair follows closely, and the filter-ID line trails by a wide
margin, with the synthetic variant worst. Two of the three predicted
weaknesses are visible right here: real audio beats synthetic by 8.7
points (B over A), and the same filters that fail as labels help as
augmentations (C). The third, the static task never rewarding temporal
sensitivity, surfaces in the reason probe below, where the only candidate
above chance is the one trained with a temporal auxiliary.

Our reading of the reconstructive advantage is that it is a scale effect:
reconstruction supervises every output cell densely, which suits a small
encoder and a small decoder, while contrastive objectives lean on batch
size and on carefully tuned augmentation families, resources that thin
out at 1.17M parameters and batch 256. We offer this as an interpretation
the numbers are consistent with, since the bake-off itself cannot prove
it; at larger scale the gap may well close, and published results with
large transformers suggest it does.

\paragraph{Reason.} The subject-wise block of Table~\ref{tab:reason} is
flat at chance. The
temporal-auxiliary candidate E is the only one above 0.5, a weak but
suggestive signal that whatever reason information exists lives in temporal
dynamics that mean-pooled embeddings mostly discard. Adaptation on real
cries, which contributes 6 to 9 points in \cite{gorin2023}, contributes
nothing here. The HuBERT control removes the capacity explanation: a model
80 times larger, pretrained on orders of magnitude more audio, lands on the
same floor and shows the same leaky-split jump. What remains is the labels:
in-the-moment parent guesses, 84\% of clips in one class, minority classes
of 8 to 25 clips.

\begin{figure}[t]
\centering
\includegraphics[width=\linewidth]{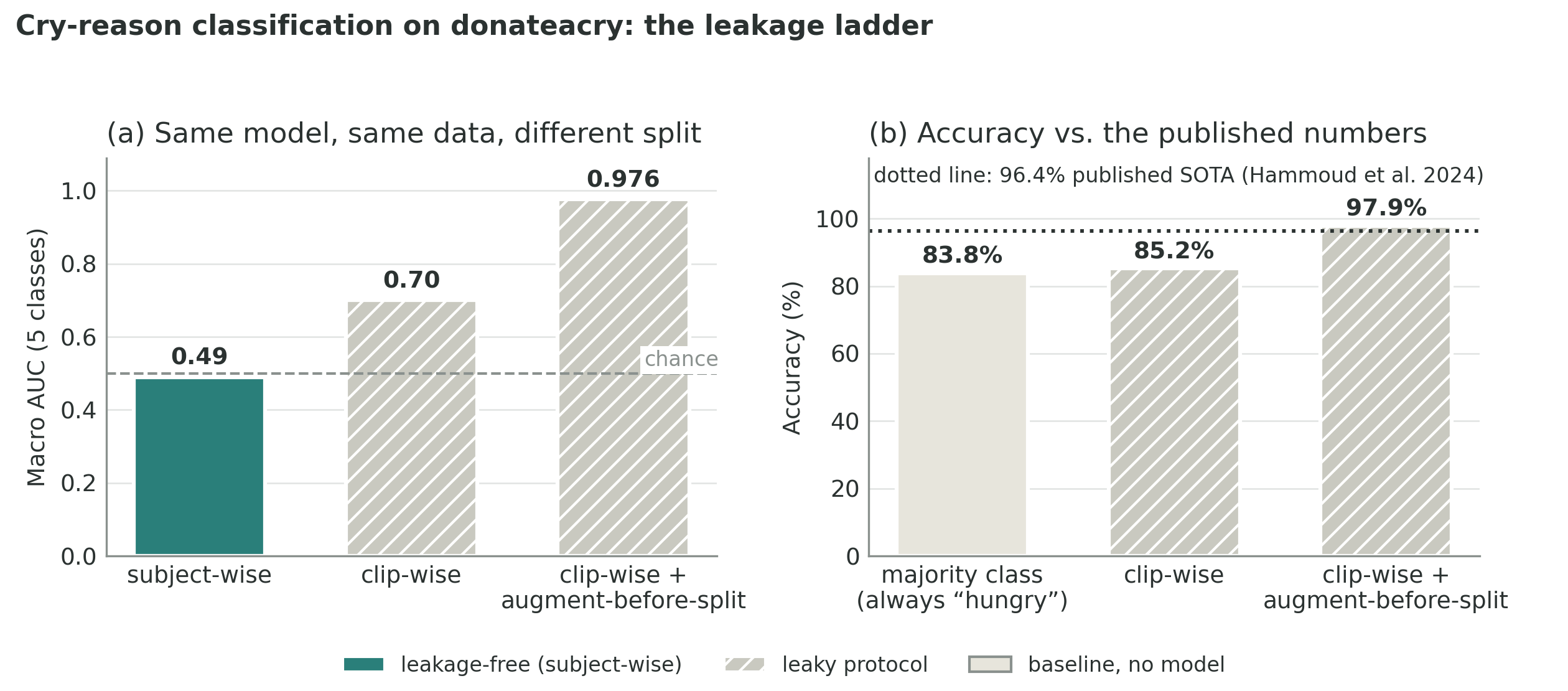}
\caption{The leakage ladder on donateacry. Left: one model, one dataset,
three split protocols; only the subject-wise protocol (solid) is valid.
Right: the leaky protocols land on the accuracy range published as state
of the art, and the majority-class baseline shows how little unbalanced
accuracy means on this corpus.}
\label{fig:leak}
\end{figure}

\paragraph{Leakage, measured end to end.} Four measurements on identical
data and models, summarized in Figure~\ref{fig:leak}, close the argument.
First, switching frozen embeddings from subject-wise to clip-wise splits
lifts the masked-spectrogram encoder from 0.38 to 0.61 AUC and HuBERT
from 0.42 to 0.61, while candidate E, already at 0.54, moves only to
0.58: subject identity alone hands a fixed representation up to 0.2 AUC,
and compresses very different encoders onto the same leaky ceiling. Second, always predicting the majority class already
yields 83.8\% accuracy on this corpus, so unbalanced accuracy is close to
meaningless here. Third, our end-to-end fine-tuning under a clip-wise
split, with no other change, reaches 85.2\% accuracy (0.70 AUC): the
network now tunes its features to individual infants. Fourth, applying
augmentation before splitting, so that synthetic variants of test clips
sit in the training set, reaches 97.9\% accuracy (0.976 AUC), matching the
96.4\% reported as state of the art on donateacry by Hammoud et
al.~\cite{frontiers2024}, whose split is described only as ``80\% training
and 20\% testing'' with no mention of subject identity. The same model,
under subject-wise evaluation, measures $0.49$ macro AUC. We conclude that
donateacry supports cry detection research while lending no support to
reason classification claims, and that reviewers in this domain should
expect three things as a matter of routine: splits by subject, a
majority-class baseline printed next to any accuracy figure, and
augmentation applied only after splitting.

\paragraph{Augmentation.} The four fine-tuning arms (real only, +noise,
+vocoder, +both) score 0.47, 0.46, 0.43 and 0.49 macro AUC, statistically
indistinguishable at these fold variances. Twenty times more labeled audio
from the same 204 infants adds no cross-subject signal. Data collection
plans in this domain should be sized in subjects.

\section{Reproducibility}

Everything ran on one consumer laptop (NVIDIA RTX 4060 Laptop, 8 GB VRAM;
16 GB system RAM, with feature caches read via memory mapping). Each
pretext pretraining takes under an hour; per-fold adaptation takes minutes
per fold; the complete study fits in a few evenings. Seeds are fixed in
the configuration files, every number in this paper is written to a CSV by
the script that produced it, and the per-clip license manifest doubles as
the exact data inventory. The pipeline is twelve numbered scripts, from
corpus download with checksums through the leakage diagnostics; the
repository in the abstract footnote contains all of it, and rebuilds the
study from the original public sources since no audio is redistributed.

\section{Limitations}

The negative result is a statement about donateacry's labels, not about the
task: \cite{gorin2023} demonstrates that with clinical labels the reason
task is learnable. Our encoders are small by design, and mean pooling over
3-second patches erases sequence structure; a latent-predictive temporal
objective in the JEPA family is the natural next candidate once
better-labeled data exists. Detection is reported as ranking quality; an
operating point in false alarms per night, and a systematic multi-SNR
evaluation, remain to be run. Finally, the human labeling pass was
performed by a single listener.

\section{Conclusion}

With one compact encoder, a fixed budget and a leakage-free protocol, six
pretext tasks sorted cleanly and informatively on cry detection:
reconstruction came first at 0.988 AUC from a linear probe with no cries
in pretraining, the contrastive pair followed closely, explicit filter
identification trailed, and every gap in the ranking traces back to a
single design ingredient. On cry reason the picture inverted, since every
candidate collapsed to chance, the collapse survived both a 94M-parameter
capacity control and 1.8 hours of domain adaptation, and adopting the
literature's own clip-wise, augment-first protocol was enough to
resurrect the published 96\%+ accuracy from the very model that had just
measured 0.49 macro AUC.

Three inexpensive practices follow for anyone working on this problem:
splitting by subject everywhere, including the self-supervised stages;
printing the majority-class baseline next to any accuracy figure, which
on donateacry stands at 83.8\% before a single parameter is trained; and
augmenting only after splitting. Under those rules donateacry remains a
legitimate resource for cry detection and a misleading one for cry
reason, and data collection plans in this domain should be sized by the
number of infants they reach before the number of hours they record.

What would move the reason task forward is equally concrete. On the data
side it means labels tied to observed outcomes, whether clinical
annotation as in~\cite{gorin2023} or retrospective resolution recorded by
caregivers, gathered across enough infants for subject-wise evaluation to
have statistical teeth. On the modeling side, the one above-chance cell
in our results points at temporal structure, so objectives that predict
latent trajectories over time are the natural next candidates for this
harness, which we release with the rest of the code. Until such data
exists, we hope the leakage ladder of Figure~\ref{fig:leak} serves as the
reference point it was built to be: the measured floor of this benchmark,
and the price of ignoring it.

\end{document}